\documentclass[sn-aps,iicol]{sn-jnl}

\usepackage{graphicx}%
\usepackage{multirow}%
\usepackage{amsmath,amssymb,amsfonts}%
\usepackage{amsthm}%
\usepackage{mathrsfs}%
\usepackage[title]{appendix}%
\usepackage{xcolor}%
\usepackage{color}
\usepackage{textcomp}%
\usepackage{manyfoot}%
\usepackage{booktabs}%
\usepackage{algorithm}%
\usepackage{algorithmicx}%
\usepackage{algpseudocode}%
\usepackage{listings}%

\usepackage{pifont}
\usepackage{times} 
\usepackage{microtype} 
\usepackage[normalem]{ulem}
\usepackage{enumitem} 
\usepackage{footmisc} 
\usepackage{rotating} 
\usepackage{siunitx} 
\usepackage{bbding}
\usepackage{xspace}
\usepackage{comment}
\usepackage{tcolorbox} 
\usepackage{color}
\usepackage{xcolor}
\usepackage{amsmath}
\usepackage{amsfonts}
\usepackage{amssymb}
\usepackage{amsthm}
\usepackage{mathrsfs}
\usepackage{bm}
\usepackage{array}
\usepackage{tabularx}
\usepackage{longtable}
\usepackage{booktabs}
\usepackage{makecell}
\usepackage{multirow}
\usepackage{colortbl} 
\usepackage{float}
\usepackage{graphicx}
\usepackage{caption}
\usepackage{wrapfig}
\usepackage{sidecap} 
\usepackage{subcaption}
\usepackage{listings}
\usepackage[title]{appendix}%
\usepackage{textcomp}%
\usepackage{manyfoot}%
\usepackage{algorithm}%
\usepackage{algorithmicx}%
\usepackage{algpseudocode}%
\usepackage{bbding}

\usepackage{hyperref}
\usepackage[capitalize]{cleveref} %

\makeatletter
\DeclareRobustCommand\onedot{\futurelet\@let@token\@onedot}
\def\@onedot{\ifx\@let@token.\else.\null\fi\xspace}

\def\eg{\emph{e.g}\onedot} 
\def\ie{\emph{i.e}\onedot}

\makeatother

\begin{document}

\title[Article Title]{Multi-Modal Prototypes for Open-World Semantic Segmentation}


\author[1]{\fnm{Yuhuan} \sur{Yang}$^*$}\email{yangyuhuan@sjtu.edu.cn}

\author[1]{\fnm{Chaofan} \sur{Ma}$^*$}\email{chaofanma@sjtu.edu.cn}

\author[1]{\fnm{Chen} \sur{Ju}}\email{ju\_chen@sjtu.edu.cn}

\author[1]{\fnm{Fei} \sur{Zhang}}\email{ferenas@sjtu.edu.cn}

\author[1,2]{\fnm{Jiangchao} \sur{Yao}$^\text{\Envelope}$}\email{Sunarker@sjtu.edu.cn}

\author[1,2]{\fnm{Ya} \sur{Zhang}}\email{ya\_zhang@sjtu.edu.cn}

\author[1,2]{\fnm{Yanfeng} \sur{Wang}$^\text{\Envelope}$}\email{wangyanfeng622@sjtu.edu.cn}

\affil[1]{\orgdiv{Cooperative Medianet Innovation Center}, \orgname{Shanghai Jiao Tong University}}

\affil[2]{
\orgdiv{Shanghai AI Laboratory}}


\abstract{
    In semantic segmentation, generalizing a visual system to both seen categories and novel categories at inference time has always been practically valuable yet challenging.
    To enable such functionality, existing methods mainly rely on either providing several support demonstrations from the visual aspect or characterizing the informative clues from the textual aspect (\eg, the class names).
    Nevertheless, both two lines neglect the complementary intrinsic of low-level visual and high-level language information, while the explorations that consider visual and textual modalities as a whole to promote predictions are still limited.
    To close this gap, we propose to encompass textual and visual clues as \emph{multi-modal prototypes} to allow more comprehensive support for open-world semantic segmentation, and build a novel prototype-based segmentation framework to realize this promise.
    To be specific, unlike the straightforward combination of bi-modal clues, we decompose the high-level language information as multi-aspect prototypes and aggregate the low-level visual information as more semantic prototypes, on basis of which, a fine-grained complementary fusion makes the multi-modal prototypes more powerful and accurate to promote the prediction.
    Based on an elastic mask prediction module that permits any number and form of prototype inputs,
    we are able to solve the zero-shot, few-shot and generalized counterpart tasks in one architecture.
    Extensive experiments on both PASCAL-$5^i$ and COCO-$20^i$ datasets show the consistent superiority of the proposed method compared with the previous state-of-the-art approaches, and a range of ablation studies thoroughly dissects each component in our framework both quantitatively and qualitatively that verify their effectiveness.
}

\keywords{multi-modality, open-world, prototype, semantic segmentation}



\maketitle

\section{Introduction}

\begin{figure*}[t]
    \begin{center}
        \includegraphics[width=\linewidth]{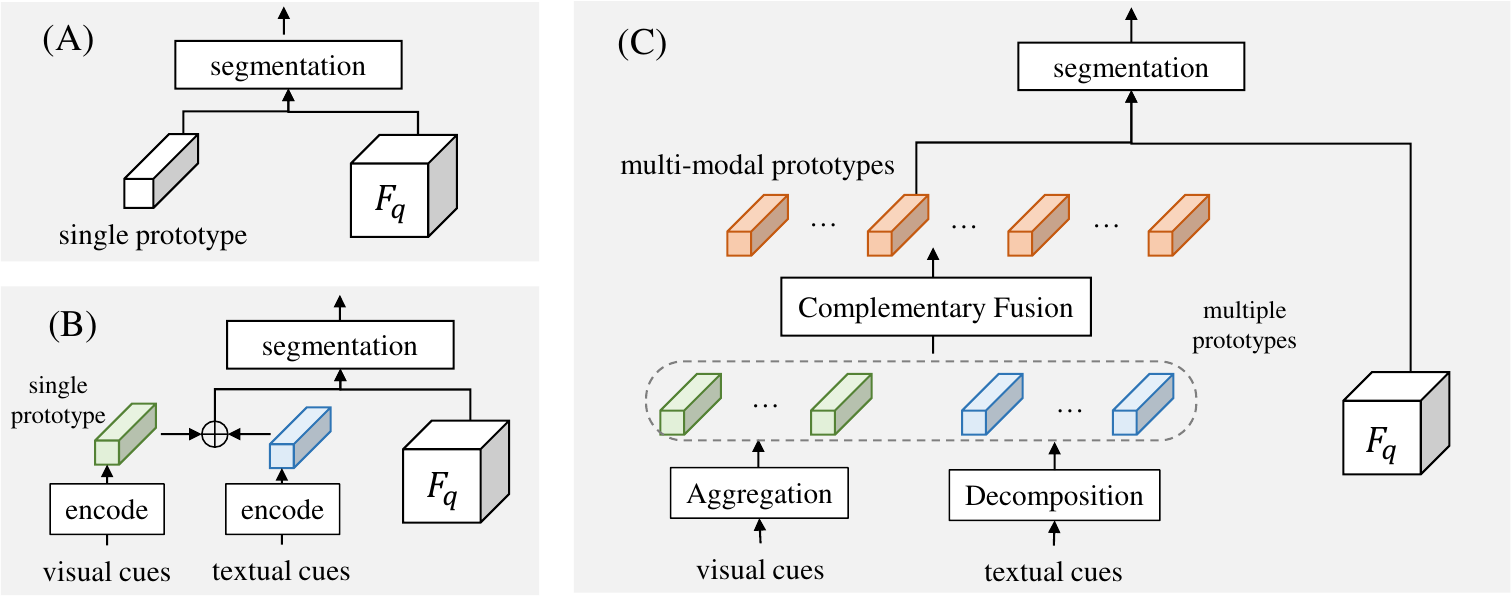}
    \end{center}
    \caption{\textbf{(A) Single-prototype-based paradigm.}
    The model learns a single prototype from uni-modal information and uses it as a semantic indicator for segmentation tasks.
    \textbf{(B) Straightforward combination.} 
    It's ineffective to straightforwardly combine the two modality through prototype addition.
    \textbf{(C) Multi-modal-prototype-based segmentation framework.} 
    Multiple prototypes are obtained through visual aggregation and textual decomposition, followed by the integration of complementary fusion to acquire multi-modal prototypes.
    }
    \label{fig:settings}
\end{figure*}

\label{sec:intro}
Semantic segmentation as one fundamental task in computer vision has made remarkable progress with the development of large-scale datasets~\citep{lin2014microsoft} and deep neural networks~\citep{DeepLab,UNet,FCN,zhao2017pyramid}.
Despite these advances, most studies~\citep{Strudel_2021_ICCV, xie2021segformer, cheng2021maskformer} focus on the \textit{closed-world} setting, where the basic categories of interest maintain the same throughout both training and inference. However,
this assumption does not always hold in practice, as the target categories are unlikely stationary, which limits the potential of early closed-world segmentation methods.

Recent explorations to address this problem yield a more challenging setting, namely, \textit{open-world} semantic segmentation, which can be roughly summarized into two lines from the visual aspect or the textual aspect:
(1) Methods based on visual cues are usually referred as one- or few-shot segmentation~\citep{CANet19,ye2021learning}, which aim to segment unseen categories with limited visual examples.
{Most of these methods aim to learn a semantic center from the given visual demonstrations~\citep{Wang_2019_ICCV,tian2020pfenet}, and then use it as pixel-wise classifiers~\citep{zhang2022FPTrans}.}
(2) Zero-shot segmentation~\citep{zs3net,xu2021}
usually utilizes the textual class information to segment seen and novel categories without any visual examples.
Specially, with the emergence of vision-language pre-training, many studies propose to exploit the aligned representations from pre-trained vision-language models by constructing textual class names as classifiers~\citep{ma2022fusioner,liang2022open,li2022languagedriven}.

{As shown in Figure~\ref{fig:settings}(A), we summarize the aforementioned two lines as single-prototype-based paradigm with the uni-modal information.
The model first learns a single vector, named as prototype, from either visual or textual modality, and then uses it as a semantic indicator for segmentation tasks.}
However, both the visual aspect and the textual aspect can be combined as these two types of clues can be complementary in terms of information granularity (perception level v.s. semantic level). 
Note that, it is ineffective to straightforwardly combine them due to the difficulty of semantic alignment in the latent space. 
{The textual and visual modality lies in two extremes: the textual modality is comprehensive and compressed, while the visual modality is intricate and plentiful.}
To close this gap, we propose a more general multi-modal-prototype-based segmentation framework shown in Figure~\ref{fig:settings}(C).
Intuitively, we propose to decompose the textual clue into fine-grained descriptions and transform them into multiple prototypes. Correspondingly, we aggregate the visual clue into different multiple prototypes to improve their semantic correspondence. On the basis of fine-grained textual and visual prototypes, we design an efficient complementary fusion module that extracts more powerful multiple multi-modal prototypes to promote the open-world semantic segmentation.

Specifically, our framework consists of four parts: visual prototype extractor, textual prototype extractor, complementary fusion module and elastic mask prediction module.
In visual prototype extractor, as visual features are intricate and plentiful across regions, we aggregate features across regions to establish multiple inherently consistent prototypes.
For textual prototype extraction, as texts like class names tend to be concise and condensed possibly with lexical ambiguity, we decompose their semantics into fine-grained descriptions automatically, and extract the corresponding prototypes.
To learn powerful multi-modal prototypes, we design a complementary fusion module that effectively mediates the relevance between prototypes of different classes and modalities.
When computing the segmentation mask, we use a class-agnostic aggregation to combine results of different levels that permits any number and form of prototype inputs.
With this design, our model can effectively handle zero-shot, few-shot and generalized few-shot tasks in one architecture.
Our contributions can be summarized into three folds:

\begin{itemize}
    \item We present a novel multi-modal framework for open-world segmentation, which effectively leverages the complementary visual and textual cues to construct more powerful multi-modal prototypes to promote the segmentation performance.

    \item We design a fine-grained multi-prototype generation and fusion mechanism that efficiently merge the information of textual modality and visual modality, and flexibly incorporate the multi-modal prototypes to promote diverse open-world segmentation tasks in one architecture.

    \item We conduct extensive experiments on two widely used datasets and achieve state-of-the-art performance in all benchmarks with various settings. Through a range of ablation studies, we prove the effectiveness of each component in our framework and provide the insights about the design.

\end{itemize}

\section{Related Work}

{

\subsection{Using Visual Clues for Segmentation}
Methods using visual clues are commonly mentioned as \textbf{few-shot segmentation} (FSS). 
They learn to segment novel categories with limited image-mask pairs as visual examples.

\subsubsection{Few-shot Segmentation}
In terms of the utilization of support information, few-shot segmentation can be broadly divided into two categories: single-vector-based methods and dense-feature-based methods.
(1).
Initially, the majority of research concentrated on feature representation learning~\citep{Wang_2019_ICCV,tian2020pfenet,liu_part-aware_2020}. 
Specifically, they aim to condense the key features of a category into a single vector.
And this vector is regarded as the semantic center of that category.
These features act either as a pixel-wise classifier~\citep{zhang2022FPTrans} or are integrated into the decoder~\citep{zhang_self-guided_2021,nguyen_feature_2019,lu_simpler_2021,NEURIPS2022_f7fef21d} to facilitate segmentation on novel categories.
(2).
Dense-feature-based methods focuses on exploiting the dense features from support images~\citep{democratic_2020,zhang_cycle_2022}. 
Techniques such as calculating pixel-to-pixel similarities using 4D convolution~\citep{min2021hypercorrelation} or employing attention-based mechanisms~\citep{hong_cost_2021} are prevalent in this area.

\subsubsection{Generalized Few-shot Segmentation} 
Most methods above are initially designed for a binary setting, targeting the segmentation of a single novel class per instance.
Research efforts like GFSS~\citep{GFSS} and DIaM~\citep{hajimiri2023diam} aim to enhance the versatility of FSS techniques.
GFSS~\citep{GFSS} extended the setting to be able to predict all potential base and novel classes. DIaM~\citep{hajimiri2023diam} proposed a baseline based on distilled information maximization loss.
Similar to GFSS, our method is also designed to handle all potential base and novel classes in one forward pass.

\subsection{Using Textual Clues for Segmentation}
Methods using textual clues mainly aim to explore the synergies between visual and textual modality by utilizing textual representations such as category name or description for segmentation.

\subsubsection{Zero-shot Segmentation}
Similar to the few-shot segmentation setting, zero-shot semantic segmentation (ZS3) targets on learning segmentation models for novel categories, but without any visual examples.
To achieve the base-to-novel mapping, they turned to language side for help.
Early works include using generative models to create visual features from word embeddings~\citep{zs3net,csrl,CaGNet} or developing a joint embedding space for pixels and semantic words~\citep{spnet,JoEm}. 
Recent studies leverage the capabilities of pre-trained vision-language models (VLMs) to enhance the alignment between images and text~\citep{li2022languagedriven,ma2022fusioner}.

\subsubsection{Open-vocabulary Segmentation}
With the rise of pre-trained vision-language models such as CLIP~\citep{Radford21}, the newly proposed \textbf{open-vocabulary segmentation} focuses on a slightly different perspective.
They aim to train a model on a set of fundamental categories such as COCO~\citep{lin2014microsoft} to learn the visual-textual alignment.
Then, they apply this knowledge to segment other no-training datasets regardless of the category overlap~\citep{Zhang_2023_ICCV,yu2023convolutions,liang2022open,DBLP:conf/nips/ZhangZLHMZY0W23}.

Further, another line of research is dedicated to training open-vocabulary segmentation models in a \textbf{weakly supervised} manner.
They do not require pixel-wise annotated segmentation datasets for training, and instead, they only use large-scale image-text datasets such as LAION-5B~\citep{schuhmann2022laion} or CC12M~\citep{changpinyo2021cc12m}.
These approaches aim to align the model's understanding between images and texts using only captioning datasets, subsequently applying this knowledge to segmentation tasks~\citep{cha2022tcl,cai2023mixreorg,xu2022groupvit,ghiasi2022scaling}.

\subsection{Prototype-based Learning}
Prototype has been early studied in bridging the gap between base and novel categories during training.
\cite{snell2017prototypical} proposed prototypical networks for few-shot and zero-shot learning. This network is based on the idea that ``there exists an embedding in which points cluster around a single prototype representation for each class''.
\cite{dong2018few} firstly introduced prototype learning in few-shot segmentation area.
Following this line of research, the majority of few-shot segmentation methods concentrate on prototype learning~\citep{Wang_2019_ICCV,tian2020pfenet,liu_part-aware_2020}. 
The prototypes act either as a pixel-wise classifier~\citep{zhang2022FPTrans} or are integrated into the decoder~\citep{zhang_self-guided_2021,nguyen_feature_2019,lu_simpler_2021,NEURIPS2022_f7fef21d} to facilitate segmentation on novel categories.

Prototypes have also been used to boost image-text alignment in language-guided segmentation tasks.
\cite{zs3net} learned to generate prototypes from text embeddings for zero-shot segmentation.
And \cite{spnet} used the word embedding directly as prototypes to project base and novel categories into the same embedding space.
With the come-up of CLIP, textual features from CLIP text encoder are widely used as prototypes to guide open-vocabulary segmentation~\citep{li2022languagedriven,liang2022open}.
}

{
    Technically, different from above methods (especially the open-vocabulary segmentation methods \textit{e.g.,} FC-CLIP and OpenSeeD \textit{etc.}), 
    out technical innovation lies on a multi-modal-prototype-based segmentation framework that unite the vision and language modalities into multi-modal prototype representation to capture complex semantics. 
    Specially, we list the following points in detail to clarify the critical differences:
    \textbf{(1)} While previous methods are mainly restricted to generate single prototype, we propose to enrich the prototypes from both visual and textual modality, via our \textit{M-splitting} and \textit{textual decomposition} pipeline;
    \textbf{(2)} We design a \textit{complementary fusion module} to foster cross-modality communication for better prototype representation, while previous methods mainly handle the two modalities separately.
    \textbf{(3)} Different from previous single-prototype-based pipeline that is only able to predict a single target, we design a \textit{multiple-prototype-based mask prediction pipeline} for more effective prototype utilization. Besides, we provide a in-depth experimental comparison in Section~\ref{sec:more-comparison} to address the concerns of readers regarding the pre-training-based or  open-vocabulary segmentation methods.
}

\begin{figure*}
    \centering
    \includegraphics[width=1\linewidth]{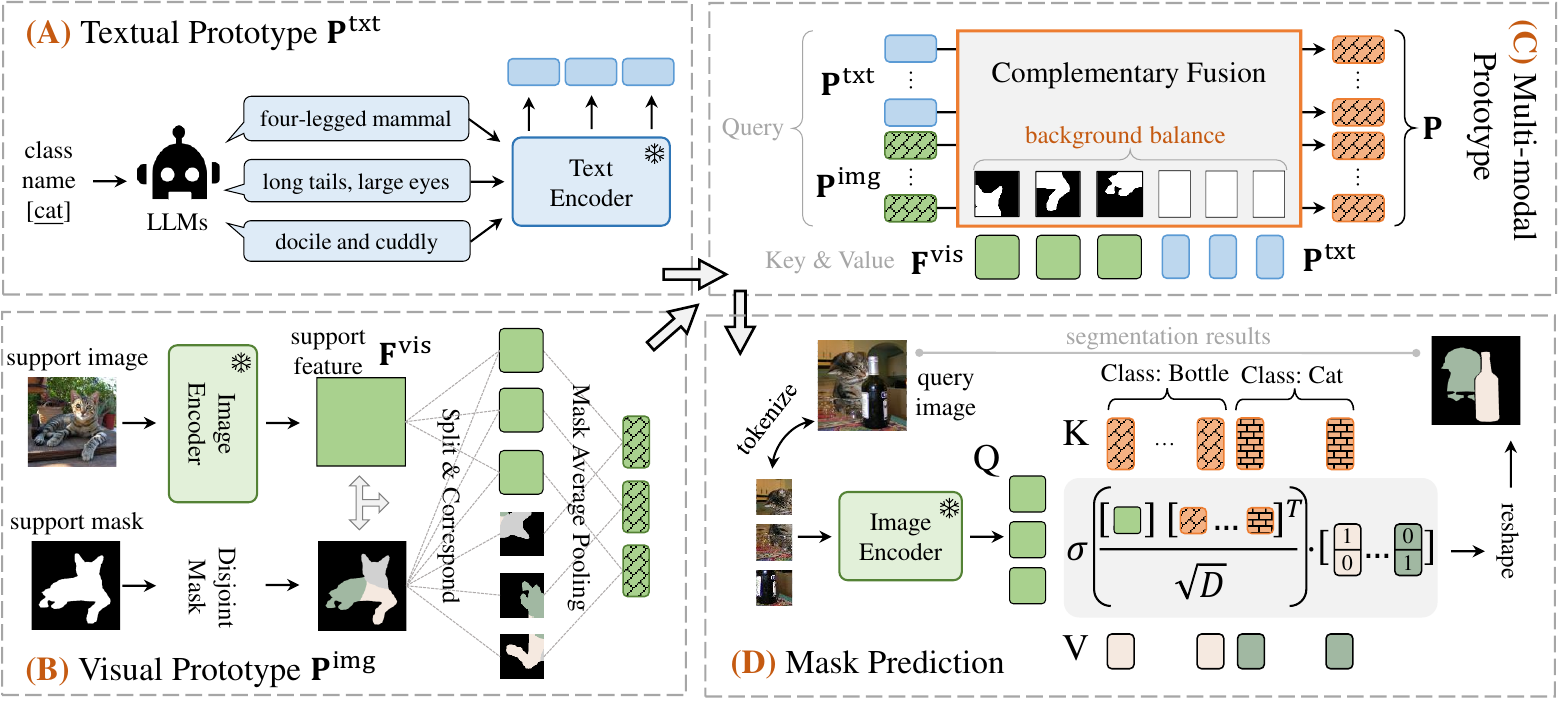}
    \caption{{\textbf{Framework Overview.}
                \textbf{(A) Textual prototypes through decomposition}: to enrich the context and eliminate ambiguity, we 
                decompose their semantics into fine-grained descriptions using LLMs.
                \textbf{(B) Visual prototypes through aggregation}: 
                we split mask into regions and aggregate features accordingly to establish multiple inherently consistent prototypes.
                \textbf{(C) Fusing multi-modal prototypes}: to learn powerful multi-modal prototypes, we design a complementary fusion module that effectively mediates the relevance between prototypes.
                \textbf{(D) Mask Prediction}: we design a comprehensive mask calculating module
                that permits any number and form of prototype inputs.}}
    \label{fig:archi}
\end{figure*}

\section{Method}
In the following, we start by introducing task setup in Section~\ref{sec:task}; and then we describe the prototype extraction from visual and textual data in Section~\ref{sec:visual} and Section~\ref{sec:textual}, followed by multi-modal fusion in Section~\ref{sec:cross-modal}; we detail mask prediction in Section~\ref{sec:mask_calc} and describe training and inference in Section~\ref{sec:loss}.

\subsection{Preliminaries}
\label{sec:task}
\subsubsection{Task Formulation}
Given an image $\mathbf{I}_q\in\mathbb{R}^{H\times W\times 3}$ with height $H$ and width $W$, open-world segmentation aims to train one model $\Phi$ to classify each pixel into $\mathcal{C}$ semantic classes:
\begin{equation}
    \mathbf{M}_q = \Phi(\mathbf{I}_q;\, \Theta) \in\{0, 1\}^{H\times W \times\mathcal{C}}.
\end{equation}
Formally, during training, image-mask pairs from seen (base) classes $\mathcal{C}_{\mathrm{seen}}$ are given, \ie, $\{(\mathbf{I}, \mathbf{M}) \sim \mathcal{C}_{\mathrm{seen}}\}$; while during testing, the model is evaluated beyond seen classes, \ie, $\{\mathbf{I} \sim \mathcal{C}_{\mathrm{seen}}\cup \mathcal{C}_{\mathrm{unseen}}\}$.

To enable open-world capability, two ways have been explored to characterize novel semantics.
The first is to leverage textual class names for unseen classes~\citep{ma2022fusioner,liang2022open,xu2021,li2022languagedriven}, \ie, $\{\mathcal{T} \sim \mathcal{C}_{\mathrm{unseen}}\}$. 
The second is to provide several image-mask (or image-bbox) support exemplars for unseen classes~\citep{tian2020pfenet,CANet19,Wang_2019_ICCV,ye2021learning}, \ie, $\{\mathcal{S} = (\mathbf{I}, \mathbf{M}) \sim \mathcal{C}_{\mathrm{unseen}}\}$. 
We here consider learning unseen semantics from support visual examples together with textual information.

\subsubsection{The Proposed Architecture}
As shown in Figure~\ref{fig:archi}, our proposed framework contains four main components, namely, one visual prototype extractor $\Phi_{\mathrm{img}}$, one textual prototype extractor $\Phi_{\mathrm{txt}}$, one multi-modal complementary fusion $\Phi_{\mathrm{fuse}}$ and one {mask prediction} $\Phi_{\mathrm{mask}}$.
For category semantics, $\Phi_{\mathrm{img}}$ takes support exemplars $\mathcal{S}=\{(\mathbf{I}_s,\mathbf{M}_s)\}$ as inputs to output visual prototype $\mathbf{P^{\mathrm{img}}}$; while $\Phi_{\mathrm{txt}}$ takes decomposed text $\mathcal{T}$ as inputs to output textual prototype $\mathbf{P^{\mathrm{txt}}}$.
After multi-model prototype fusion, 
we use $\Phi_{\mathrm{mask}}$ to output masks of the test (query) image $\mathbf{I}_q$.
The above pipeline can be summarized as
\begin{align}
    \begin{split}
        & \mathbf{P^{\mathrm{img}}} = \Phi_{\mathrm{img}}(\mathbf{I}_s, \mathbf{M}_s), ~\mathbf{P^{\mathrm{txt}}} = \Phi_{\mathrm{txt}}(\mathcal{T}), \\
        & \mathbf{P} = \Phi_{\mathrm{fuse}}(\mathbf{P^{\mathrm{img}}},\mathbf{P^{\mathrm{txt}}}),
        ~\hat{\mathbf{M}}_q = \Phi_{\mathrm{mask}}\left(\mathbf{I}_q,\mathbf{P}\right).
    \end{split}
\end{align}

\subsection{{Visual Prototype Extractor}}
\label{sec:visual}
\subsubsection{Limitation of Single Prototype}
Given the support demonstration that exhibits a high similarity to query images, we aim to extract visual features as prototypes from support examples to promote semantic segmentation.
A naive way is extracting single prototype from support example through weighted sum on visual features that characterize foreground regions.
Concretely, with the visual feature $\mathbf{F}^{\mathrm{vis}}\in\mathbb{R}^{H\times W\times D}$ of the support image output by an image encoder, and the support mask $\mathbf{M}^c$ of class $c$, the single prototype can be computed as
\begin{equation}
        \mathbf{P}_{\mathrm{single}} = \Psi_{\mathrm{single}}(\mathbf{F}^{\mathrm{vis}}, \mathbf{M}^c) = \sum \omega  \mathbf{F}^{\mathrm{vis}} \in\mathbb{R}^D
\end{equation}
where $\omega = \mathbf{M}^c / (\sum\mathbf{M}^c)$.
However, one prototype may not be able to sufficiently reflect all variations \textit{w.r.t.} object of interests,
as the visual appearance of the support image may vary across different regions. 
For instance, when representing a ``tree'', the upper portion is typically characterized by a green color and dense foliage, whereas the lower part tends to exhibit a brown hue and a prominent main branch.
Mixing these two parts together may lead to a distorted representation. This results in a representation neither similar to the upper nor the lower part, thus constrains segmentation.

\begin{figure}[!t]
    \centering
    \includegraphics[width=0.8\linewidth]{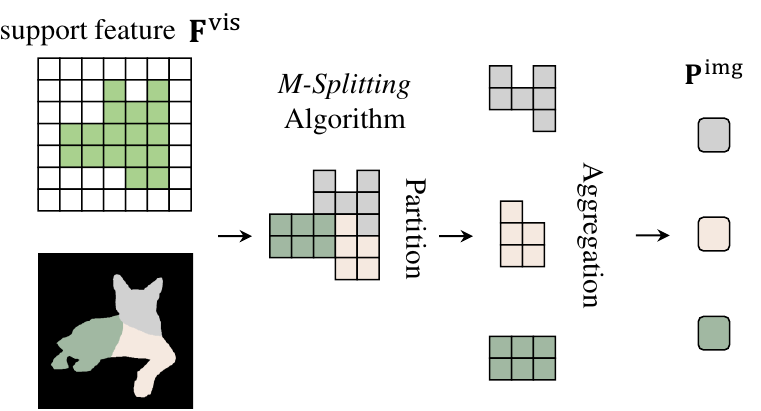}
    \caption{{\textbf{Visual prototypes through aggregation.}
                We split the support mask into several regions using the \textit{M-Splitting} algorithm and average visual feature on each region, forming several tokens as visual prototypes.}}
    \label{fig:vision_archi}
\end{figure}

\subsubsection{Elaborate Multiple Visual Prototypes}
\label{sec:multi_visual_p}
To alleviate representation distortions, we propose to aggregate different regions of an object separately as different prototypes. 
This process gives out several visual prototypes encapsulating diverse visual appearances, while conveying the same semantic meaning. Formally, we split  $\mathbf{M}^c$ of class $c$ into $n$ non-overlapping regions: $\{\mathbf{M}^c_1, \mathbf{M}^c_2,..., \mathbf{M}^c_n\}$, and then compute multiple visual prototypes with the concatenation as follows,
\begin{equation}
    \begin{aligned}
    \mathbf{P}^{\mathrm{img}}_c & = \Psi_{\mathrm{multi}}(\mathbf{F}^{\mathrm{vis}}, \mathbf{M}^c)  \\
      & = [\Psi_{\mathrm{single}}(\mathbf{F}^{\mathrm{vis}},\mathbf{M}^c_i)]_{i=1}^n \in\mathbb{R}^{n\times D},
    \end{aligned}
    \label{eq:multi_visual_P}
\end{equation}
where $[\cdot]_{i=1}^n $ will concatenate each element embedding.

\noindent\textbf{$\textbf{M}$-splitting}: To effectively divide any mask ({$\mathbf{M}^c$ in Eq~\ref{eq:multi_visual_P}, simplified as $\mathbf{M}$ in Algorithm~\ref{algo:part}}) into non-overlapping regions, we design a method named \textit{M-Splitting} inspired by Voronoi diagrams~\citep{aurenhammer1991voronoi}.
Specifically, our algorithm begins by randomly selecting an initial semantic center and iteratively chooses the farthest pixel from the known ones as the new center. With a greedy strategy, \textit{M-Splitting} provides a solution that is both fast and easy to implement, whose details are summarized in Algorithm~\ref{algo:part}. We will compare \textit{M-Splitting} with K-means in Section~\ref{exp:split}.
\begin{algorithm}
    \caption{M-Splitting algorithm}
    {
    \begin{enumerate}
        \item For a binary mask $\mathbf{M}\in\mathbb{R}^{H\times W}$, we want to split it into $n$ parts, and each part is a binary mask $\mathbf{M}_k\in\mathbb{R}^{H\times W}$, $k=1,2,\cdots,n$.
        \item Randomly select an initial point $(x_1,y_1)$, ensuring that $\mathbf{M}[x_1,y_1]=1$. Here $\mathbf{M}[x_1,y_1]$ means the value of the mask $\mathbf{M}$ at the point $(x_1,y_1)$. Add $(x_1,y_1)$ to the collection of partition centers $\mathcal{P}=\{(x_1,y_1)\}$.
        \item To select the next partition center, for an arbitrary point $(x,y)$, we define the following distance $\mathcal{D}(x,y)$. It represents the distance between $(x,y)$ and the partition center $\mathcal{P}$.
        \begin{equation}
            \mathcal{D}(x,y) = \min_{(x_i,y_i)\in\mathcal{P}}\left((x-x_i)^2+(y-y_i)^2\right)\notag
        \end{equation}
        \item Find the next point $(x^*,y^*)$ that $\mathbf{M}[x^*,y^*]=1$ and $\mathcal{D}(x^*,y^*)$ is the maximum among all the points in the mask $\mathbf{M}$.
        \item Add $(x^*,y^*)$ to the collection of partition centers $\mathcal{P}=\mathcal{P}\cup\{(x^*,y^*)\}$.
        \item Repeat steps 3-5 until $|\mathcal{P}|=n$.
        \item For each partition center $(x_k,y_k)\in\mathcal{P}$, we define a binary mask $\mathbf{M}_k$ as follows:
        \begin{equation}
            \mathbf{M}_k[x,y] = \begin{cases} 1,&
                \begin{aligned}
                    &\, \text{if } \mathcal{D}(x,y) \text{ is equal to }\\
                    &\,  (x-x_k)^2+(y-y_k)^2
                \end{aligned} \\
                \\
                0,  &\text{otherwise.}
            \end{cases}\notag
        \end{equation}
        This means, for each point $(x,y)$, we assign it to the partition center $(x_k,y_k)$ if the distance between $(x,y)$ and $(x_k,y_k)$ is the minimum among all the partition centers.
    \end{enumerate}
}
    \label{algo:part}
\end{algorithm}

\subsection{{Textual Prototype Extractor}}
\label{sec:textual}

\subsubsection{Limitation of Class Names} 
On the textual side, the typical way to acquire the prototype is by feeding the given text information into a pre-trained text encoder. In many recent studies~\citep{li2022languagedriven,ma2022fusioner},
vanilla class names with fixed prompts are directly used. However, this usually results in prototypes with the ambiguous discrimination, as class names are low informative even with  biased information.
(1) \textit{Lexical ambiguity}. A class name usually consists of one or two words, {\eg}, ``crane'' can refer to either a bird or a machine.
(2) \textit{Lexical weak-tie.} Sometimes the connection between class name and its literal meaning is weak. For instance, there is little visual similarity between Pomeranian dog and Pomerania location. Rather, Pomeranian dogs are typically differentiated based on their fox-like faces and small size. Therefore, it can be ineffective to only leverage class names in the textual prototype extraction.

\subsubsection{Decomposed Granular Descriptions}
To strengthen the representation power, 
we decompose the vanilla class names into detailed descriptions from different aspects to transform the high-level language as more granular information.
For example,
replacing ``Pomeranian dogs'' with ``dog with fox-like faces, thick and fluffy fur'' will increase the discriminating power of prototypes generated by text encoder.
Specifically, instead of hand-crafted decomposition,
we utilize large language models (LLMs)~\citep{gpt3,openai2023gpt4} to implement this goal, for their remarkable performance in semantic understanding and text generation. 
To automate this procedure with LLMs $\Phi_{\mathrm{llm}}$,
a robust querying instruction that can induce well-organized answers while filtering out irrelevant descriptions is designed. For clarity, we present
an exemplar prompt
``\texttt{What are some \underline{visual features} for distinguishing Pomeranian dogs \underline{in an image}? \underline{list} by items}''
and its response
\begin{itemize}
    \item \underline{Size}: Pomeranians are small dogs which typically weighing between 3-7 pounds ...
    \item \underline{Coat}: 
    Pomeranians have a thick, fluffy double coat of fur that comes in a variety of colors.
    \item \underline{Behavior}: Pomeranian dogs are known for ...
\end{itemize}
By parsing the answer using $\bullet$ as indicator, we can get a list of descriptions for Pomeranian dogs from different views.
Let $\mathcal{T}$ denote the class name and the above process with LLMs generates
\begin{equation}
    \mathcal{T}^{\mathrm{txt}}= \{\mathcal{T}^{\mathrm{txt}}_1, \mathcal{T}^{\mathrm{txt}}_2, ...,  \mathcal{T}^{\mathrm{txt}}_{n-1} \} = \Phi_{\mathrm{llm}}(\mathcal{T}), \nonumber
\end{equation}
where $\mathcal{T}^{\mathrm{txt}}_1, \mathcal{T}^{\mathrm{txt}}_2, ...\mathcal{T}^{\mathrm{txt}}_{n-1}$ are $n-1$ decomposed granular descriptions depicting the current class from different {perspectives}. 
By asking LLMs and parsing the answers, we can get a list of context descriptions $\mathcal{T}^{\mathrm{txt}}$ for each class. Then, we combine $\mathcal{T}^{\mathrm{txt}}$ with vanilla class names, feed into CLIP~\citep{Radford21} for text-modal embeddings. 
The resulting representation $\mathbf{P}^{\mathrm{txt}}_c \in\mathbb{R}^{n\times D}$ are denoted in the following equation.
\begin{equation}
    \mathbf{P}^{\mathrm{txt}}_c =[\Psi_{\mathrm{CLIP}}(\mathcal{T}),\Psi_{\mathrm{CLIP}}(\mathcal{T}_1^{\mathrm{txt}}),...,\Psi_{\mathrm{CLIP}}(\mathcal{T}_{n-1}^{\mathrm{txt}})],
\end{equation}
where $\Psi_{\mathrm{CLIP}}(\cdot)$ means CLIP text encoder. Note that, we maintain the class name as one independent prototype and add $n-1$ decomposed granular prototypes.

\subsection{Multi-Modal Prototype Generation%
}
\label{sec:cross-modal}
To leverage the complementary characteristic of the visual and textual prototypes and extract more powerful multi-modal counterpart, we design a complementary fusion module to realize the bi-modal knowledge alignment, interaction, and fusion with $\mathbf{P}^{\mathrm{img}}_c$ and $\mathbf{P}^{\mathrm{txt}}_c$.

\noindent\textbf{Complementary Fusion}: In this module, we employ the cross-attention mechanism to achieve our goal.
Concretely, we set \textit{query} as $\mathbf{Q} = [\mathbf{P}^{\mathrm{txt}}_c, \mathbf{P}^{\mathrm{img}}_c]\in\mathbb{R}^{2n\times D}$ ($n$ is the prototype number of each modality).
For \textit{key} and \textit{value}, 
as it is critical to incorporate the background information of the image to promote the recognition of false negatives,
we set as {$\mathbf{K}=\mathbf{V} = [\mathbf{P}^{\mathrm{txt}}_c,\mathbf{F}^{\mathrm{vis}}]\in\mathbb{R}^{(n+H\times W)\times d}$}, where
$\mathbf{F}^{\mathrm{vis}}$ is the global image feature that is acquired by forwarding the image into visual encoder and contains both foreground and background information.
Furthermore, as the background feature is involved, the cost is how to balance its importance during fusion. Here, we design a learnable weighted mask on the background portion of the attention to properly avoid its overwhelmed effect.
Thus, our multi-modal prototypes are acquired by the following balanced attention
\begin{equation}
    \begin{aligned}
        \mathbf{P}_c & = \Phi_{\mathrm{fuse}}(\mathbf{P}^{\mathrm{txt}}_c, \mathbf{P}^{\mathrm{img}}_c)                                                            \\
                   & = \sigma\left(\frac{\mathbf{Q}\mathbf{K}^\mathsf{T}}{\sqrt{D}}-\alpha\cdot \overline{\mathbf{M}}_c\right)\cdot \mathbf{V}\in\mathbb{R}^{2n\times D}.
    \end{aligned}
\end{equation}
{Here we get $\overline{\mathbf{M}}_c$ by repeating $[\mathbf{0},1-\mathbf{M}_s]\in\mathbb{R}^{n+H\times W}$ for $2n$ times along the first dimension, namely, $\overline{\mathbf{M}}_c\in\mathbb{R}^{2n\times (n+H\times W)}$. Therefore, $\overline{\mathbf{M}}_c$ gets non-zero values only on features representing background patches.}
$\alpha$ is the learnable parameter for foreground-background balance. When $\alpha\to\infty$, the output is only relevant with foreground features, and no background features are involved.

\begin{figure*}
    \centering
    \includegraphics[width=.95\linewidth]{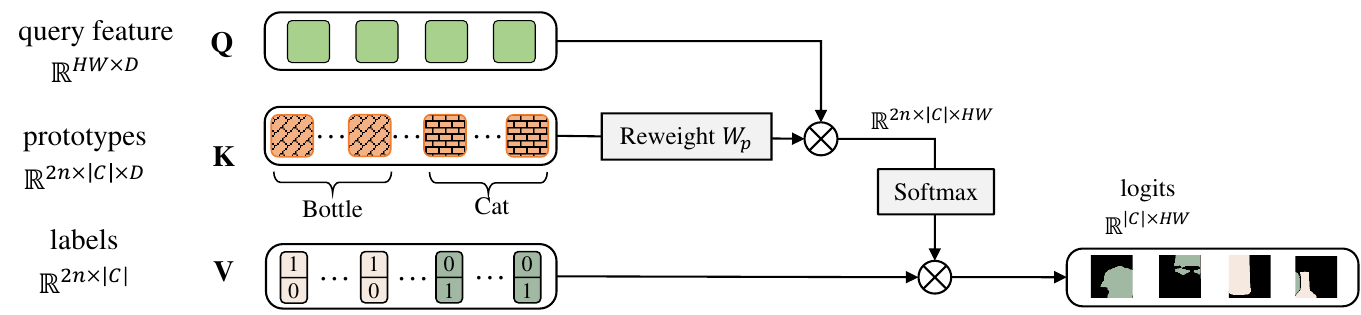}
    \caption{{\textbf{Multiple-prototype-based mask prediction pipeline.} Different prototypes are seen as independent classifiers, and compete with each other through attention mechanism. Prototypes of the same class share the same $\mathbf{V}$ and can be grouped together during the attention process.}}
    \label{fig:mask_calc}
\end{figure*}

\subsection{{
Elastic Mask Prediction}}
\label{sec:mask_calc}

\begin{algorithm}[!t]
    \caption{Attention based mask prediction}
    {
    \begin{algorithmic}[1]
        \Ensure Prototypes $\tilde{\mathbf{P}}\in\mathbb{R}^{2n\times|\mathcal{C}|\times D}$, query feature $\mathbf{F}_q\in\mathbb{R}^{HW\times D}$, label $l_p\in\{0,1\}^{2n\times|\mathcal{C}|}$, weight $W_p\in\mathbb{R}^{2n}$.

        \State $K = \tilde{\mathbf{P}}\odot W_p\in\mathbb{R}^{2n\times |\mathcal{C}|\times D}$
        \State $A=K(\mathbf{F}_q)^T\in\mathbb{R}^{2n\times |\mathcal{C}|\times HW}$
        \State $A'=$\texttt{softmax}($A/\sqrt{d}$, dim=$2n\times|\mathcal{C}|$)
        \State $p=A'\odot l_p\in\mathbb{R}^{2n\times|\mathcal{C}|\times HW}$
        \State $\hat{\mathbf{y}}=\sum_{2n,|\mathcal{C}|}p \in \mathbb{R}^{HW}$
    \end{algorithmic}}
    \label{algo:mask_att}
\end{algorithm}

In this module, we aim to get mask predictions using the set of multiple multi-modal prototypes for all $|\mathcal{C}|$ classes. 
Intuitively, we consider the multiple prototypes as independent sub-class classifiers, where each prototype competes with others for logits prediction. 
This competition is achieved through an attention mechanism, where the prototypes act as $\mathbf{K}$ and the query feature acts as $\mathbf{Q}$. 
Prototypes belonging to the same class share the same $\mathbf{V}$ and can be grouped together during the attention process. Formally, we denote $\tilde{\mathbf{P}}\in\mathbb{R}^{2n|\mathcal{C}|\times D}$ the overall prototypes that concatenate the prototypes $\mathbf{P}_c\in\mathbb{R}^{2n\times D}$ for class $c$, written as follows
\begin{equation}
    \tilde{\mathbf{P}} = [\mathbf{P}_1,\mathbf{P}_2,...,\mathbf{P}_{|\mathcal{C}|}]\in\mathbb{R}^{2n|\mathcal{C}|\times D}. \nonumber
\end{equation}
Given query feature $\mathbf{F}_q\in\mathbb{R}^{HW\times D}$, the mask prediction logits can be computed by the follow attention
\begin{equation}
    \begin{aligned}
        \hat{y} & = \sigma\left(\frac{\mathbf{Q}\mathbf{K}^\mathsf{T}}{\sqrt{D}}\right)\cdot \mathbf{V}                                                    \\
                & = \sigma\left(\frac{\mathbf{F}_q (W_p\cdot\tilde{\mathbf{P}})^\mathsf{T}}{\sqrt{D}}\right)\cdot l_P\in\mathbb{R}^{HW\times|\mathcal{C}|}.
    \end{aligned}
\end{equation}
Here $l_P\in\{0,1\}^{2n|\mathcal{C}|}$ is the one-hot label for prototypes $\tilde{\mathbf{P}}$.
To account for the varying contributions of the $2n$ prototypes, we introduce a learnable weight $W_p\in\mathbb{R}^{2n}$ to balance the importance of each prototype. 
A detailed pseudo-code is shown in Algorithm~\ref{algo:mask_att} and the  mask prediction process is illustrated in Figure~\ref{fig:mask_calc}.

\begin{figure*}[htbp]
    \centering
    \includegraphics[width=\textwidth]{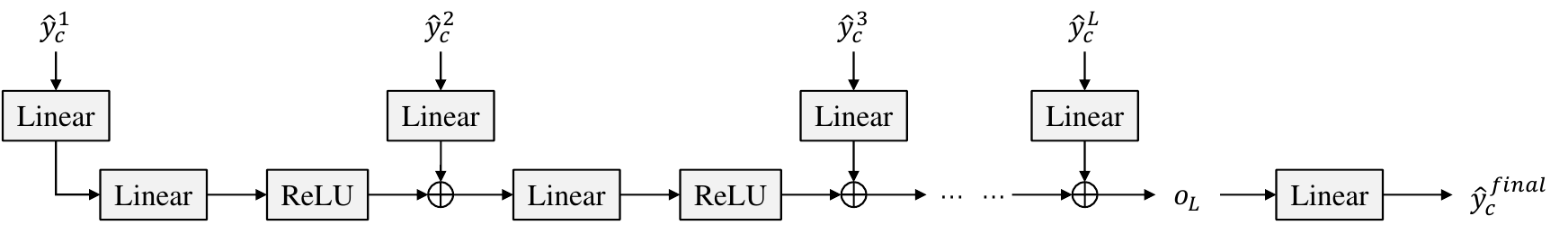}
        \caption{{\textbf{An illustration of \cref{eq:multi_level}}. The multi-level prediction is fused one-by-one to get the final prediction.}}
        \label{fig:8}
    \end{figure*}

\noindent\textbf{Multi-Level Fusion}:
In practice, as objects usually vary in scale, multi-level visual modeling is necessary, given that the last-layer output from the visual encoder may lose detailed information.
Here we introduce a multi-level fusion module to address this issue.
Specifically, we consider totally $L$ intermediate layers to form multi-level feature pyramid.
For each level, we generate a coarse mask prediction, having
$\{\hat{\mathbf{y}}^l\}_{l=1}^L$.
Then, we utilize a residual structure with skip connections to fuse them together for final mask prediction.
We formulate the fusion process in the following equation.
\begin{equation}
    \begin{aligned}
        \mathbf{o}_1 & = \mathrm{ReLU}(W_1\cdot W_{in}\hat{\mathbf{y}}_c^1+\mathbf{b}_1),                     \\
        \mathbf{o}_2 & = \mathrm{ReLU}(W_2\cdot \mathbf{o}_1 + \mathbf{b}_2) + W_{in}\hat{\mathbf{y}}_c^2,    \\
        ...          & = ...                                                                                 \\
        \mathbf{o}_L & = \mathrm{ReLU}(W_L\cdot \mathbf{o}_{L-1}+\mathbf{b}_L) + W_{in}\hat{\mathbf{y}}_c^L.
    \end{aligned}
    \label{eq:multi_level}
\end{equation}
Here, $W_{in}\in\mathbb{R}^{1\times d}$ projects the one-dimensional logits into deep feature. $W_i\in\mathbb{R}^{d\times d}$ and $\mathbf{b}_i\in\mathbb{R}^d$ are weights and bias for each level.
The final prediction is computed by $\hat{\mathbf{y}}_c^{\mathrm{final}} = W_{out}\cdot \mathbf{o}_L$, where $W_{out}\in\mathbb{R}^{d\times d}$. 
{We also provide an illustration of \cref{eq:multi_level} in \cref{fig:8}.}

\subsection{Training, Inference and Beyond}
\label{sec:loss}
During training, we have the pixel-wise annotations $\mathbf{M}_q\in\{0,1\}^{HW\times |\mathcal{C}_\mathrm{seen}|}$ on seen classes $\mathcal{C}_\mathrm{seen}$ as supervision. To optimize the proposed framework, we use the cross-entropy loss $\mathcal{L}_{\mathrm{CE}}$ over final logits $\hat{\mathbf{y}}^{\mathrm{final}}$ and those intermediate logits $\hat{\mathbf{y}}^{l}$ before fusion as follows
\begin{equation}
    \mathcal{L}_{\mathrm{all}} = \mathcal{L}_{\mathrm{CE}}(\hat{\mathbf{y}}^{\mathrm{final}}, \mathbf{M}_q)+\lambda\sum_{l=1}^{L}\mathcal{L}_{\mathrm{CE}}(\hat{\mathbf{y}}^{l}, \mathbf{M}_q),
\end{equation}
where $\lambda$ is a balancing ratio. Note that, throughout training, CLIP encoders and LLMs are frozen to reduce the computational burden, preserve the prior knowledge and avoid overfitting to the seen classes. Regarding inference, we first calculate multi-modal prototypes for unseen classes using visual demonstrations and textual data, and then concatenate them together with prototypes of seen classes for segmentation. The other pipeline remains the same as training.

\noindent\textbf{Discussion}: There has been relatively limited exploration to utilize both visual and textual cues for open-world semantic segmentation. 
One closely related work is CLIPSeg~\citep{clipseg}, which
uses FiLM~\citep{FiLM} modulated by a single CLIP's visual or textual embedding as cues for zero/one-shot segmentation.
While exhibited some initial performance, 
this approach confines the input to a single and uni-modal prototype, where prototypes are reluctantly to communicate through a straightforward linear interpolation.
On one hand, this restricts its flexibility, as it cannot handle multiple input cues (\ie, an image with a sentence), multi-class, (\ie, several semantic classes), and multi-shot (\ie, few images).
What's worse, one single prototype can potentially lose information when the semantic concept is sophisticated, causing a bottleneck in the segmentation performance.
On the contrary, 
our design introduces the concept of \textit{multiple prototypes} incorporated with \textit{multiple modalities}.
It allows us to deal with multi-modal/shot/class settings by adjusting the number of prototypes for practical use. We can also encompass a more comprehensive range of information from diverse perspectives utilizing multiple prototypes. The experiments in the following will confirm the promise of our design.

\section{Experiments}

\subsection{Datasets}
Here, we evaluate our proposed method on two prevailing benchmarks, \ie, PASCAL-$5^i$~\citep{shaban2017oneshot} and COCO-$20^i$~\citep{shaban2017oneshot}. PASCAL-$5^i$~\citep{shaban2017oneshot} is built
from PASCAL VOC 2012~\citep{everingham2015pascal}, and we follow the dataset split in~\citep{shaban2017oneshot} to evenly
split 20 object categories into four folds, each of which could be treated as unseen classes if the rest act as the training sets (seen classes). Similarly, COCO-$20^i$~\citep{shaban2017oneshot}, extracted from  MSCOCO~\citep{lin2014microsoft}, is evenly split into four folds, each with 20 classes.
Following~\citep{hajimiri2023diam}, we use all available query images for PASCAL-$5^i$ and sample $10k$ images for COCO-$20^i$.

\subsection{Task Setup}
We evaluate our method under two settings: zero/few shot (Z/FS) and generalized few shot (GFS), 
respectively (as shown in Table~\ref{table:task}). {Z/FS} aims to evaluate the model of transferring knowledge from the seen classes to the unseen ones. 
The model accepts support information of a specific class $\{(\mathbf{I},\mathbf{M}_c)\}$, and predicts a binary segmentation mask $\mathbf{M}\in\mathbf{R}^{H\times W}$ for the given query image. During training $c\in\mathcal{C}_{\mathrm{seen}}$ and during testing $c\in\mathcal{C}_{\mathrm{unseen}}$.
Based on the number of the support image-mask pairs provided, {Z/FS} is further divided into 1-shot (one image-mask pair provided), 5-shot (five pairs provided), and zero-shot (merely text information provided). 
Furthermore, we evaluate our method under the more challenging GFS setting, which is a multi-class version of {FS}. The model, instead of making a binary prediction for the query object, exerts the predicted results for all candidate classes.
Formally, the model outputs multi-class logits $\mathbf{y}\in\mathbb{R}^{H\times W\times|\mathcal{C}|}$, where $|\mathcal{C}|$ contains both the seen and unseen classes.
To sum up, {Z/FS} and GFS share same training data, but are implemented differently in following aspects:
\begin{itemize}
    \item \textbf{Classes of interest}: Z/FS focuses on unseen classes, but GFS targets on seen and unseen ones.
    \item \textbf{Model input}: Z/FS can only take support information of one class $c$, $(\mathbf{I},\mathbf{M}_c)$ where $\mathbf{M}_c$ is the binary segmentation mask for class $c$. While GFS takes a set of support information of all candidate classes, $\{(\mathbf{I}, \mathbf{M}_c),c\in\mathcal{C}\}$.
    \item \textbf{Model output}: With different input, Z/FS is only able to predict a binary mask $\mathbf{M}_c\in\{0,1\}^{H\times W}$ for a class $c$, while GFS is able to predict mask $\mathbf{M}\in\{0,1\}^{H\times W\times\mathcal{C}}$ for all classes at once.
    \item \textbf{Missing classes}: Z/FS ignores the evaluation of the missing classes, while GFS evaluates such classes by generating the predicted mask of missing class $c$ as 0.
\end{itemize}

\begin{table*}[t!]
    \centering
    \caption{{\textbf{Comparison between zero/few shot (Z/FS) and generalized few shot (GFS).} Z/FS is the setting used in CLIPSeg~\citep{clipseg}. GFS is a more difficult setting than Z/FS.}}
    \label{table:task}
    \begin{tabular}{ccc}
        \toprule
        \textbf{Setting}   & Z/FS                                           & GFS                                                            \\\midrule
        Training           & \multicolumn{2}{c}{Using the same training data}                                                                     \\\midrule
        Evaluation Classes & $\mathcal{C}_{\mathrm{unseen}}$                & $\mathcal{C}_{\mathrm{seen}}\bigcup\mathcal{C}_{\mathrm{unseen}}$ \\\midrule
        Model Input        & \makecell[c]{Information for one specific class $c$                                                                     \\$\mathcal{T}_c\ /\ (\mathbf{I}, \mathbf{M}_c)$}& \makecell[c]{Information for all classes\\$\{(\mathbf{I}, \mathbf{M}_c),c\in\mathcal{C}\}$}\\\midrule
        Model Prediction   & \makecell[c]{Binary mask for the given class $c$                                                                      \\$M\in\{0,1\}^{H\times W}$} & \makecell[c]{Multi-dimensional mask for each class\\$M\in\{0,1\}^{H\times W\times\mathcal{C}}$}\\\midrule
        Missing classes    & Ignore                                         & Predicted as $\mathbf{0}$                                         \\
        \bottomrule
    \end{tabular}
\end{table*}

\subsection{Evaluation Metrics}
We use the mean Intersection over Union (mIoU) as the evaluation metric. This  is a widely used evaluation metric for segmentation tasks.
Here we average classes from $\mathcal{C}_{\mathrm{seen}}$ and $\mathcal{C}_{\mathrm{unseen}}$ separately. Formally:
\begin{equation}
    \begin{aligned}
        &\textbf{Seen} = \frac{1}{|\mathcal{C}_\mathrm{seen}|}\sum_{c\in\mathcal{C}_\mathrm{seen}}\mathrm{IoU}_c,\\
        &\textbf{UnSeen} = \frac{1}{|\mathcal{C}_\mathrm{unseen}|}\sum_{c\in\mathcal{C}_\mathrm{unseen}}\mathrm{IoU}_c,
    \end{aligned}\notag
\end{equation}
where $\mathrm{IoU}_c$ is the IoU for class c.
Following~\citep{ye2021learning} we also report the harmonic mean (\textbf{HIoU}) of \textbf{Seen} and \textbf{UnSeen} to show the model's generalizability.
\begin{equation}
    \textbf{HIoU} = \frac{2\times \textbf{Seen}\times \textbf{UnSeen}}{\textbf{Seen}+\textbf{UnSeen}}.\notag
\end{equation}

\begin{table*}[!t]
    \centering
    \vspace{-1em}
    \caption{\textbf{Comparison with SOTA methods under Z/FS setting on COCO-$20^i$ (top-half results) and PASCAL-$5^i$ dataset (bottom-half results).} We provide the baselines using either textual or visual information. Our method outperforms all baselines under each modality. Our full model (multi-modal) further enhances the performance.}
    \begin{tabular}{cc|cccc|c}
        \toprule
        Method                                & Input for Unseen Classes             & Fold-0   & Fold-1   & Fold-2   & Fold-3   & Mean     \\\midrule
        ZS3~\citep{zs3net}                    & \multirow{4}{*}{\makecell[c]{Textual                                                        \\class names \& descriptions\\(ZS setting)}} & 18.8     & 20.1     & 24.8     & 20.5     & 21.1     \\
        LSeg~\citep{li2022languagedriven}     &                                      & 22.1     & 25.1     & 24.9     & 21.5     & 23.4     \\
        Fusioner~\citep{ma2022fusioner}       &                                      & 23.6     & 28.2     & 26.2     & 24.1     & 25.5     \\
        \bf Ours (Text Only)                  &                                      & \bf 26.5 & \bf 30.8 & \bf 26.3 & \bf 24.1 & \bf 26.9 \\
        \midrule
        PPNet~\citep{liu_part-aware_2020}     & \multirow{13}{*}{\makecell[c]{Visual                                                        \\example images \& masks\\(FS setting)}}                                      & 28.1    & 30.8    & 29.5    & 27.7    & 29.0       \\
        PFENet~\citep{tian2020pfenet}         &                                      & 36.5     & 38.6     & 34.5     & 33.8     & 35.8     \\
        RePRI~\citep{REPRI}                   &                                      & 32.0     & 38.7     & 32.7     & 33.1     & 34.1     \\
        CAPL~\citep{GFSS}                     &                                      & -        & -        & -        & -        & 39.8     \\
        VAT~\citep{hong_cost_2021}            &                                      & 39.0     & 43.8     & 42.6     & 39.7     & 41.3     \\
        HSNet~\citep{min2021hypercorrelation} &                                      & 36.3     & 43.1     & 38.7     & 38.7     & 39.2     \\
        CWT~\citep{lu_simpler_2021}           &                                      & 32.2     & 36.0     & 31.6     & 31.6     & 32.9     \\
        CyCTR~\citep{zhang_cycle_2022}        &                                      & 38.9     & 43.0     & 39.6     & 39.8     & 40.3     \\
        NTRENet~\citep{liu2022NTRENet}        &                                      & 36.8     & 42.6     & 39.9     & 37.9     & 39.3     \\
        SSP~\citep{fan2022ssp}                &                                      & 35.5     & 39.6     & 37.9     & 36.7     & 37.4     \\
        RPMG-FSS~\citep{zhang2023Rpmg}        &                                      & 38.3     & 41.4     & 39.6     & 35.9     & 38.8     \\

        \bf Ours (Image Only)                 &                                      & \bf 42.0 & \bf 45.1 & \bf 44.6 & \bf 41.9 & \bf 43.4 \\
        \midrule
        CLIPSeg~\citep{clipseg}               & \multirow{2}{*}{\makecell[c]{Textual \& Visual}}   & -        & -        & -        & -        & 33.3     \\
        \bf Ours (Full)                       &                                      & \bf 42.4 & \bf 48.5 & \bf 46.3 & \bf 45.5 & \bf 45.7 \\
        \bottomrule
        \\
        \toprule
        ZS3~\citep{zs3net}                    & \multirow{3}{*}{\makecell[c]{Textual                                                        \\class names \& descriptions\\(ZS setting)}}& 40.8 & 39.4 & 39.3 & 33.6 & 38.3\\
        LSeg~\citep{li2022languagedriven}     &                                      & 52.8     & 53.8     & 44.4     & 38.5     & 47.4     \\
        Fusioner~\citep{ma2022fusioner}       &                                      & 46.8     & 56       & 42.2     & 40.7     & 46.4     \\
        \midrule
        PPNet~\citep{liu_part-aware_2020}     & \multirow{10}{*}{\makecell[c]{Visual                                                        \\example images \& masks\\(FS setting)}}& 48.6 & 60.6 & 55.7 & 46.5 & 52.8\\
        PFENet~\citep{tian2020pfenet}         &                                      & 61.7     & 69.5     & 55.4     & 56.3     & 60.8     \\
        RePRI~\citep{REPRI}                   &                                      & 59.8     & 68.3     & 62.1     & 48.5     & 59.7     \\
        CAPL~\citep{GFSS}                     &                                      & -        & -        & -        & -        & 62.2     \\
        HSNet~\citep{min2021hypercorrelation} &                                      & 64.3     & 70.7     & 60.3     & 60.5     & 64.0     \\
        CWT~\citep{lu_simpler_2021}           &                                      & 56.3     & 62.0     & 59.9     & 47.2     & 56.4     \\
        CyCTR~\citep{zhang_cycle_2022}        &                                      & 65.7     & 71.0     & 59.5     & 59.7     & 64.0     \\
        NTRENet~\citep{liu2022NTRENet}        &                                      & 65.4     & 72.3     & 59.4     & 59.8     & 64.2     \\
        RPMG-FSS~\citep{zhang2023Rpmg}        &                                      & 63.0     & 73.3     & 56.8     & 57.2     & 62.6     \\
        \bf Ours (Image Only)                 &                                      & 66.3     & 72.1     & 58.9     & 58.4     & \bf 63.9 \\
        \midrule
        CLIPSeg~\citep{clipseg}               & \multirow{2}{*}{\makecell[c]{Textual \& Visual}}   & -        & -        & -        & -        & 59.5     \\
        \bf Ours (Full)                       &                                      & \bf 68.0 & \bf 73.5 & \bf 60.1 & \bf 60.5 & \bf 65.5 \\
        \bottomrule
    \end{tabular}%
    \label{tab:single_modal}
\end{table*}

\begin{table*}[!t]
    \centering
    \caption{{\textbf{Comparison with SOTA methods under generalized few shot (GFS) setting on PASCAL-$5^i$.} We report mean results over all 5 folds. \textbf{HIoU} is the harmonic mean of \textbf{Seen} and \textbf{UnSeen}. Our method achieves the best \textbf{HIoU}, and a significant improvement on \textbf{UnSeen} categories, indicating the strong generalization to novel categories.}}
    \begin{tabular}{c|ccc|ccc}
        \toprule
                                           & \multicolumn{3}{c|}{1-Shot} & \multicolumn{3}{c}{5-Shot}                                                                      \\
        Method                             & \textbf{Seen}               & \textbf{UnSeen}            & \textbf{HIoU}  & \textbf{Seen}  & \textbf{UnSeen} & \textbf{HIoU}  \\\midrule
        CANet~\citep{CANet19}              & 8.73                        & 2.42                       & 3.79           & 9.05           & 1.52            & 5.29           \\
        PANet~\citep{Wang_2019_ICCV}       & 31.88                       & 11.25                      & 16.63          & 32.95          & 15.25           & 24.1           \\
        PFENet~\citep{tian2020pfenet}      & 8.32                        & 2.67                       & 4.04           & 8.83           & 1.89            & 5.36           \\
        SCL~\citep{zhang_self-guided_2021} & 8.88                        & 2.44                       & 3.38           & 9.11           & 1.83            & 5.47           \\
        RePRI~\citep{REPRI}                & 20.76                       & 10.50                      & 13.95          & 34.06          & 20.98           & 27.52          \\
        CAPL~\citep{GFSS}                  & 64.80                       & 17.46                      & 27.51          & 65.43          & 24.43           & 44.93          \\
        BAM~\citep{BAM}                    & 71.60                       & 27.49                      & 39.73          & 71.60          & 28.96           & 50.28          \\
        DIaM~\citep{hajimiri2023diam}      & 70.89                       & 35.11                      & 46.96          & 70.85          & 55.31           & 63.08          \\
        \textbf{Ours}                               & \textbf{71.71}              & \textbf{39.44}             & \textbf{50.89} & \textbf{72.22} & \textbf{57.53}  & \textbf{64.04} \\
        \bottomrule
    \end{tabular}
    \label{table:all-main}
\end{table*}

\subsection{Baselines}

We include most of the recent FS works  ~\cite{min2021hypercorrelation, lu_simpler_2021, zhang_cycle_2022}  as the baselines for a sound comparison.
For methods in ZS, we include LSeg~\citep{li2022languagedriven} and Fusioner~\citep{ma2022fusioner} as recent representatives, where only textual information for unseen classes are provided.
CLIPSeg~\citep{clipseg} is the only multi-modal method that has been evaluated under Z/FS setting.
To ensure a fair comparison, we also provided the single-modal versions of our method.

Theoretically, an FS-evaluated model can be extended to GFS setting by conducting inference on each candidate class and aggregating the results through voting for final predictions.
However, this proves impractical in real-world applications due to its substantial computational demands. For instance, in the case of COCO-$20^i$ with 80 candidate classes, an FS model would necessitate evaluating 80 times on each image, resulting in an extremely time-consuming process.
Therefore, under GFS setting, we only included baselines that can be evaluated efficiently. For example, \citep{GFSS} assumes that only the last classification layer needs to be modified for different classes, allowing most of the model's forward process to be calculated only once. Similarly, \citep{BAM} and \cite{hajimiri2023diam} can also be evaluated in parallel with only minor modifications.

\subsection{Implementation Details}
For a fair comparison, all experiments are conducted on CLIP ResNet50 backbone. As ResNet encoder uses progressive downsampling, we extract all inner features smaller than $1/4$ resolution (feature pyramid level $l\in\{1,2,3\}$). Features with the same resolution share the same model parameter. We set the number of prototypes for single modality $n=3$, and loss weight $\lambda=0.01$. We treat the ``background'' as a special class in $\mathcal{C}_{seen}$. SGD optimizer is used with learning rate $lr=1e-3$.
    {For LLMs, we use OpenAI API on calling \texttt{gpt-4}~\citep{openai2023gpt4} model.}
All experiments are conducted using 2 NVIDIA RTX A6000 GPUs.

\subsection{Performance compared with SOTAs}

\noindent\textbf{Comparison Under Z/FS Setting.}
Table~\ref{tab:single_modal} presents the results obtained under Z/FS setting using both PASCAL-$5^i$ and COCO-$20^i$ dataset. As can be seen, the single-modal version of our method has already demonstrated impressive performance, surpassing all baselines. Besides, the multi-modal version has the potential to further enhance this performance, showcasing the best results overall.

\begin{table}[!t]
    \centering
    \caption{{\textbf{Comparison with SOTA methods under generalized few shot (GFS) setting on COCO-$20^i$.} The evaluation is conducted under 1-shot setting. We report the mean results over all 5 folds. \textbf{HIoU} is the harmonic mean of \textbf{Seen} and \textbf{UnSeen}.}}
    \begin{tabular}{c|ccc}
        \toprule
                                      & \multicolumn{3}{c}{1-Shot}                                    \\
        Method                        & \textbf{Seen}              & \textbf{UnSeen} & \textbf{HIoU}  \\\midrule
        RePRI~\citep{REPRI}           & 5.62                       & 4.74            & 5.14           \\
        CAPL~\citep{GFSS}             & 43.21                      & 7.21            & 12.36          \\
        BAM~\citep{BAM}               & {49.84}             & 14.16           & 22.05          \\
        DIaM~\citep{hajimiri2023diam} & 48.28                      & 17.22           & 25.39          \\
        \textbf{Ours}                          & \bf 49.86                      & \textbf{19.48}  & \textbf{28.02} \\
        \bottomrule
    \end{tabular}
    \label{table:coco-main}
\end{table}

\noindent\textbf{Comparison Under GFS Setting.}
\label{sec:gfss}
Table~\ref{table:all-main} and Table~\ref{table:coco-main} shows the results compared under GFS setting using PASCAL-$5^i$ and COCO-$20^i$ respectively.
Our method demonstrates superior performance compared to the state-of-the-art DIaM by achieving a $4.33\%$ improvement on the \textbf{UnSeen} metric in the 1-shot setting on the PASCAL-$5^i$ dataset and $2.26\%$ on the COCO-$20^i$ dataset. These results highlight the efficacy of incorporating text as guidance, as it enables better generalization to novel categories.
Moreover, our method achieves the highest \textbf{HIoU} score, indicating that our model excels in accurately segmenting both seen and unseen classes simultaneously.

\subsection{Ablation Studies}

\noindent\textbf{Mask Splitting Algorithm. }
\label{exp:split}
Comparing our proposed \textit{M-Splitting} with K-means, both algorithms can split a mask into different regions. However, K-means requires multiple iterations to find the optimal clusters, while \textit{M-Splitting} combines random and greedy strategies. This allows our method to be much faster than K-means in terms of computational speed.

Table~\ref{table:kmean} demonstrates the significant difference in speed between these two algorithms.
Since K-means is an iterative algorithm, here we fix the maximum number of iterations (n\_iter=3 or 10) to make the total complexity more controllable.
We report the total inference time on a randomly picked set of 10 images.
It can be observed that even K-means is fast for a few numbers of iterations (n\_iter=3), \textit{M-splitting} exhibits a significant advantage in terms of computational time.

\begin{figure}
    \centering
    \includegraphics[width=\linewidth]{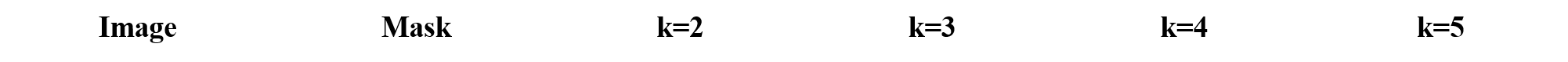}
    \includegraphics[width=\linewidth]{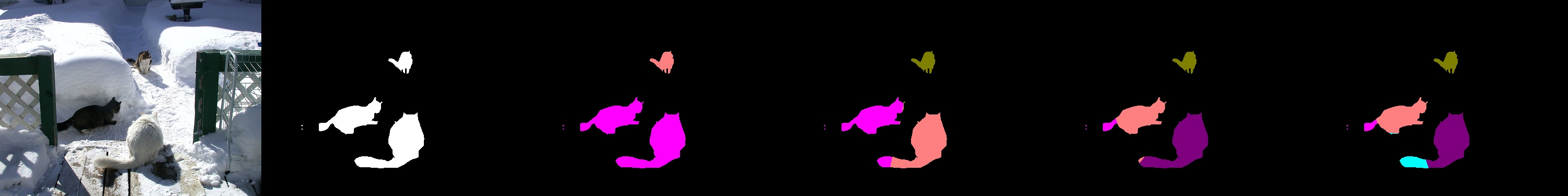}
    \vspace{1ex}
    \includegraphics[width=\linewidth]{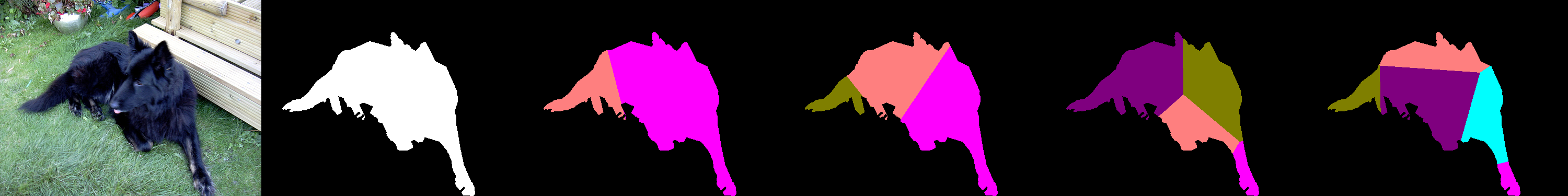}
    \includegraphics[width=\linewidth]{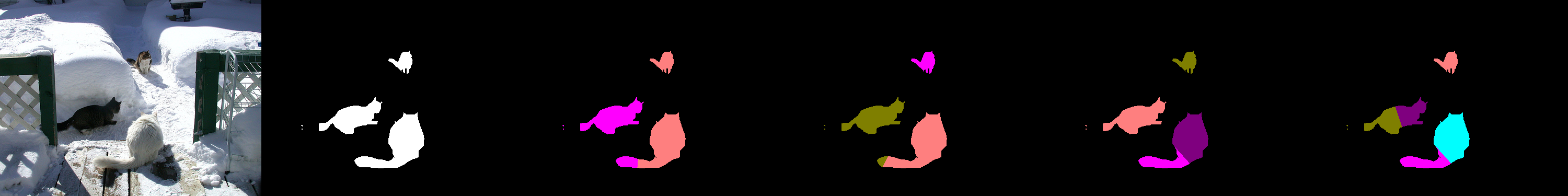}
    \includegraphics[width=\linewidth]{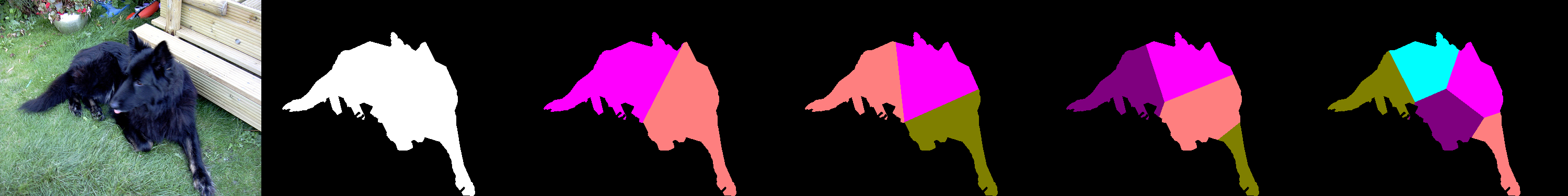}
    \caption{\textbf{Mask splitting result using \textit{M-splitting} (top-half) and K-means (bottom-half)}. The two algorithms are both able to split mask into reasonable regions.}
    \label{fig:kmean}
\end{figure}

Figure~\ref{fig:kmean} shows the resulting masks of  \textit{M-splitting} (up) and  K-means (down) under different $k$. It can be observed that both two algorithms perform reasonably. Taking time and space efficiency into consideration, we choose to use \textit{M-splitting} for prototype extraction.

\begin{table}
    \centering
    \caption{{\textbf{Total inference time cost by different algorithms.} We report the time for applying each algorithm. ``n\_iter'' is the maximum number of iterations for K-means. \textit{M-splitting} is greatly ($\sim$ 200 times) faster.}}
    \begin{tabular}{ccc}
        \toprule
         \textit{M-splitting} & K-means (n\_iter=10) & K-means (n\_iter=3) \\\midrule
        \textbf{0.043} s / image     & 22.9 s / image                & 8.36 s / image              \\
        \bottomrule
    \end{tabular}
    \label{table:kmean}
\end{table}

\noindent\textbf{Mask Splitting Number. }
There exists a trade-off between the number of masks split $n$ and the training and inference speed.
A larger $n$ corresponds to more prototypes, which means the model can capture details of the support image better. In the extreme case, each pixel of the support image can be represented as a prototype. However, an excessive number of prototypes can increase the model's size and complicate training. Therefore, we need to find a balance between the two.

Table~\ref{table:num_vpt} gives the results of different $n$.
According to the performance, increasing $n$ from 1 to 5 shows a steady improvement on results. 
This demonstrates the effectiveness of multiple prototypes.
However, larger values of $n$ become impractical under the GFS setting, particularly when dealing with numerous classes.
\begin{table} [!t]
    \centering
    \centering
    \caption{{\textbf{Effects of the number of visual prototypes.} 
    We observe a consistent improvement in performance as the number of visual prototypes increases.
However, larger values of $n$ become impractical under the GFS setting, particularly when dealing with numerous classes.
    }}
    \begin{tabular}{cccc}
        \toprule
        {Visual Prototype Number ($n$)} & \textbf{Seen}  & \textbf{UnSeen} & \textbf{HIoU}  \\\midrule
        {1    }                         & 68.82          & 27.59           & 39.39          \\
        {3    }                         & 69.81          & 32.31           & 44.17          \\
        {5    }                         & \textbf{70.21} & \textbf{33.53}  & \textbf{45.39} \\
        \bottomrule
    \end{tabular}
    \label{table:num_vpt}
\end{table}

\noindent\textbf{Description Number. }
Here, we study the impact of the number of decomposed descriptions on the performance of our model. The results, presented in Table~\ref{table:num_tpt}, are reported for $n\in{1,3,5}$.
It is important to note that $n=1$ represents vanilla class names without any decomposition. 
Comparing the results, we observe that decomposing class names leads to improved performance, highlighting the effectiveness of our textual decomposition design.
However, increasing the value of $n$ also introduces additional noise, which can have side effects. As a result, the performance of our model decreases when $n=5$ compared to $n=3$.

\begin{table}[!t]
    \centering
    \centering
    \caption{{\textbf{Ablation on the number of descriptions.} Decomposing class names into multiple descriptions increases discriminative information and achieves higher score. However, too much information may introduce noise.}}
    \begin{tabular}{cccc}
        \toprule
        {Descriptions Number} & \textbf{Seen}  & \textbf{UnSeen} & \textbf{HIoU}  \\\midrule
        {1    }               & 69.21          & 30.53           & 42.37          \\
        {3    }               & \textbf{69.81} & \textbf{32.31}  & \textbf{44.17} \\
        {5    }               & 70.32          & 30.85           & 42.89          \\
        \bottomrule
    \end{tabular}
    \label{table:num_tpt}
\end{table}

\noindent\textbf{Multi-layer Fusion.}
As shown in Section \ref{sec:mask_calc}, we employ a multi-layer fusion strategy to combine the outcomes obtained from a multi-level feature pyramid.
Here, we examine the contributions of each level to the final result. Specifically, we refer to the feature from the deepest level of the backbone as $L=1$, while the shallower levels are denoted as $L=2$ and $L=3$, respectively.
Table \ref{tab:multi_layer} presents the results achieved by utilizing each level independently, as well as combinations of different levels.

The result reveals that the last level ($L=1$) contributes the most to the final result. This is due to the deeper feature containing richer semantic information, and also aligns better with the textual description. Including shallower features ($L=2$) improves the segmentation result by incorporating low-level visual features like texture and color. However, the addition of even shallower features ($L=3$) does not yield significant improvements on \textbf{UnSeen} results. Consequently, we conclude the fusion process at this point and disregard features with $L>3$.

\begin{table}[!t]
    \centering
    \centering
    \caption{{
    \textbf{Contribution of feature pyramid levels to final result.} 
     $L=1$ contributes the most, as it provides richer semantic information and better alignment with the textual description.
    The inclusion of $L=2$ enhances the result by incorporating low-level visual features such as texture and color.
    }}
    \begin{tabular}{ccc|ccc}
        \toprule
        $L=1$      & $L=2$      & $L=3$      & \bf Seen  & \bf UnSeen & \bf HIoU  \\\midrule
        -          & -          & \checkmark & 3.06      & 1.52       & 2.03      \\
        -          & \checkmark & -          & 12.27     & 10.03      & 11.04     \\
        \checkmark & -          & -          & 44.33     & 24.41      & 31.48     \\
        \checkmark & \checkmark & -          & 57.75     & 30.58      & 39.99     \\
        \checkmark & \checkmark & \checkmark & \bf 69.81 & \bf 32.31  & \bf 44.17 \\
        \bottomrule
    \end{tabular}
    \footnotetext{Here $L=1$ refers to the deepest level of the backbone.}
    \label{tab:multi_layer}
    \vspace{-2em}
\end{table}
\begin{table}[t!]
    \centering
    \centering
    \setlength{\tabcolsep}{0.6em}
    \vspace{-1em}
    \captionof{table}{{\textbf{Ablation study on provided information.} \ding{172}-\ding{174} for textual information, \ding{175}-\ding{176} ablate the visual part. Both visual and textual information contributes to the final result.}}
    \begin{tabular}{ccccc|ccc}
        \toprule
                   & \multicolumn{2}{c}{Visual} & \multicolumn{2}{c|}{Textual}                                                                 \\
                   & Img                        & Anno                         & Name                                 & Desc          & \textbf{Seen} & \textbf{UnSeen} & \textbf{HIoU}        \\\midrule
        \ding{172} & \checkmark                 & mask                         & \checkmark                           & \checkmark    & \textbf{69.81} & \textbf{32.31}  & \textbf{44.17} \\
        \ding{173} & \checkmark                 & mask                         & \checkmark                           & -             & 68.91 & 30.59  & 42.37          \\
        \ding{174} & \checkmark                 & mask                         & -                                    & -             & 69.10 & 27.09  & 38.92          \\\midrule
        \ding{175} & \checkmark                 & box                          & \checkmark                           & \checkmark    & 62.39 & 24.32  & 35.00          \\
        \ding{176} & -                          & -                            & \checkmark                           & \checkmark    & 61.58 & 21.50  & 31.87
        \\
        \bottomrule
    \end{tabular}%
    \label{table:support_info}
\end{table}

\begin{table*}[!t]
    \caption{Comparison on ``Fantastic Beats'' dataset.}
    \small
    \begin{tabular}{ccccccc}
        \toprule
        &FC-CLIP&SEEM&LISA&Ours (Zero-Shot)&Ours (One-Shot)\\
        \midrule
        Training Dataset & \makecell{COCO-Panoptic\\(130 categories)} & \makecell{COCO-Panoptic, \\ RefCOCO, RefCOCO+, \\ G-Ref, LVIS} & Hybrid dataset$^\text{[a]}$ &\multicolumn{2}{c}{COCO (60 categories subset)}\\
        IoU & 50.9 & 52.1 &  60.1 & 51.6 & 78.5 \\
        \bottomrule
    \end{tabular}
    \small{[a] It contains ADE20K, COCO-Stuff, PACO-LVIS, PartImageNet, PASCAL-Part, RefCOCO, RefCOCO+, G-Ref, LLaVA-150k and its own ReasonSeg dataset.}
    \label{tab:fb_comparison}
\end{table*}

\noindent\textbf{Robustness to the Missing Information. }
Our method requires detailed description from both visual and textual modality. Here we systematically weaken the information provided in each modality, to investigate on the situation where these descriptions are weakened, or missing.
For textual side, we remove class description (\ding{173}).
For the visual side, we weaken the support information, by replacing mask annotations with bounding-boxes (\ding{175}).
We have also tested on single-modal situation (\ding{174}, \ding{176}).

Table~\ref{table:support_info} shows our ablation results. Clearly, experiment \ding{172} with full support information achieves the highest performance. Comparing \ding{173} and \ding{174} with \ding{172} we find that enriching textual context with detailed information could largely improve the performance. Besides, pixel-wise annotation for the support images is also important, since it provides an accurate indication of what a foreground should look like. Relaxing this constraint to box annotations will surely introduce noise into vision prototype representation (\ding{175}).
Finally, our model showed an acceptable performance even when all image information was absent (\ding{176}), suggesting that our model is robust in the extreme scenarios.

\begin{figure}[!t]
    \centering
    \begin{minipage}[c]{.55\linewidth}
        \centering
        \includegraphics[width=0.92\textwidth]{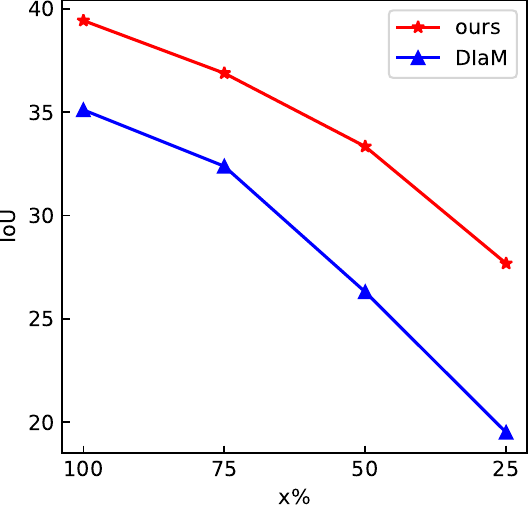}
    \end{minipage}\hfill
    \begin{minipage}[c]{.43\linewidth}
        \makeatletter\def\@captype{figure}
        \caption{\textbf{Model robustness to inaccurate information.} The annotation mask is gradually eroded to 75\%, 50\% and 25\% of the original size. Our model is more robust to incomplete masks.}
    \label{fig:robust}
    \end{minipage}
\end{figure}

\noindent\textbf{Robustness to the Inaccurate Information. }
{When producing visual prototypes in Sec~\ref{sec:multi_visual_p}, the mask $\mathbf{M}$ is essential for the model to be aware of the interested category. 
For instance, in images containing multiple object categories such as cats and dogs, $\mathbf{M}$ is used to isolate the category under study (\eg, cats) by masking out the others.
This process underscores the importance of $\mathbf{M}$ in directing the model's focus, which is particularly vital when the input image encompasses multiple object categories.}
Here to see the model's robustness, we conducted further investigations on the situation where the provided mask is inaccurate.
Specifically, we gradually eroded the edges of the annotation mask until its area was reduced to 75\%, 50\%, and 25\% of the original size.
The results are shown in Figure~\ref{fig:robust}. It clearly demonstrates that compared with DIaM, our model exhibits greater resilience to incomplete masks.

\begin{table*}[htbp]
  \centering
  \caption{Comparison with Open-Vocabulary Methods.}
  \label{tab:compare_openvoc}
  \begin{tabular}{ccc|ccc}
    \toprule
    Method & Backbone & Supervision & A-150 & PC-59 & PA-21 \\
    \midrule
    OpenSeeD~\citep{Zhang_2023_ICCV} & Swin-L &\multirow{4}{*}{\makecell{Cross-dataset transfer\\(via text)}} & 23.4&-&- \\
    OVSeg~\citep{liang2022open} & ResNet-101 && 24.8&53.3&- \\
    FC-CLIP~\citep{yu2023convolutions} & ResNet-50 && 23.3&50.5&75.9\\
    Ours & ResNet-50 && \textbf{25.3}&\textbf{54.6}&\textbf{77.5} \\
    \midrule
    Ours & ResNet-50 &\makecell{Cross-dataset transfer\\(via text + vision)}& \textbf{30.2}&\textbf{61.0}&\textbf{83.2} \\
    \textcolor{gray}{LISA~\citep{lai2023lisa}} & \textcolor{gray}{SAM ViT-H} & \textcolor{gray}{\makecell{Fully supervised}} & \textcolor{gray}{57.4} & \textcolor{gray}{70.5} & \textcolor{gray}{85.4}\\
    \bottomrule
  \end{tabular}
\end{table*}

{

\subsection{Compared with Other Powerful Segmentation models}\label{sec:more-comparison}
Currently, a new type of segmentation models, which we summarized as ``\textit{general segmentation architectures}'' has been proposed~\citep{zou2024segment,lai2023lisa,qi2023aims}.
General segmentation architectures aim to unite all segmentation task formulation, and give a powerful model supporting a range of human-computer interactions.
They are able to segment ``everything'' by absorbing nearly all public available datasets into training. 
However, despite great success, they're still not able to accept image and text simultaneously as target representation. 
What's worse, they lack the ability to deal with newly arisen categories and concepts.

\begin{figure*}[t!]
    \centering
    \includegraphics[width=1\linewidth]{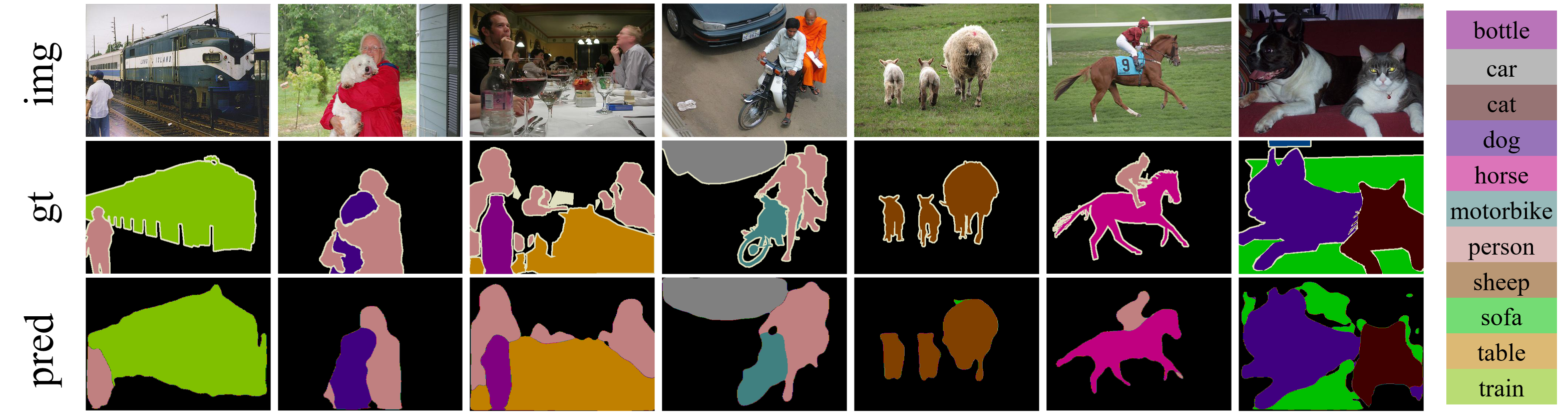}
    \caption{{\textbf{Visualization of 5-Shot Results From PASCAL-$5^i$.} As can be seen, with \texttt{bus,car,cat,chair,cow} as unseen classes, our model can segment both seen and unseen classes well.}}
    \label{fig:vis}
\end{figure*}

To further address this problem and make a fair comparison, we here provide an additional evaluation result to compare with these general segmentation architectures as well as the SOTA open-vocabulary segmentation model on a newly proposed dataset.
The dataset named ``Fantastic Beats'' is proposed by AttrSeg~\citep{ma2023attrseg}.
It collects 20 fantastic beats appeared in Harry Potter movie. Since these movies appeared after 2021, they are not included in neither the training data of general segmentation architectures nor ours, and could be considered as totally UnSeen categories without the worry of data leaking problem.
We provide a transfer evaluation on such a dataset.

In Table~\ref{tab:fb_comparison}, we compare our method with open-vocabulary method FC-CLIP~\citep{yu2023convolutions} and general segmentation architectures SEEM~\citep{zou2024segment} and LISA~\citep{lai2023lisa}.
``Zero-Shot'' means we only input textual descriptions as indicator, and ``One-Shot'' means we input both textual descriptions and image examples. As can seen, we can achieve a comparable performance with FC-CLIP with just less than half of the training categories.
And when image examples are provided, our method is able to perform greatly better than all the other methods.
This demonstrates the great power of our multi-modal prototypes in segmenting novel categories even with very restricted training data.
Note that, we propose our method not to compare with, but to further boost these general segmentation architectures. 
We provide a new perspective on addressing and understanding the ``base-to-novel'' mapping problem, and a more effective way on human-computer interaction.

}

{
\noindent\textbf{Compared with Open-vocabulary methods. }
In Table~\ref{tab:compare_openvoc}, we also compare our method with previous open-vocabulary methods FC-CLIP~\citep{yu2023convolutions}, OVSeg~\citep{liang2022open} and OpenSeeD~\citep{Zhang_2023_ICCV} on ADE20k (A-150), PASCAL-Context (PC-59) and PASCAL VOC (PA-21) datasets. We follow FC-CLIP's setting and use its ResNet-50 baseline for fair comparison.
Note that, for ``\textit{general segmentation architectures}'' such as LISA, they have already been \textit{trained} on these datasets. 
That is to say, there is no actual novel class for LISA during the test phase,  and it is not fairly comparable with open-vocabulary methods.
Thus, we regard LISA here as a fully supervised reference, and roughly considered it as an upper bound of other open-vocabulary methods.

As shown in Table~\ref{tab:compare_openvoc}, our method outperforms all open-vocabulary methods. 
Besides, using only a smaller backbone (ResNet-50), our method is even comparable with fully supervised upper-bound on PA-21 dataset by introducing vision information.

}

\begin{figure*}[t!]
    \centering
    \includegraphics[width=1\linewidth]{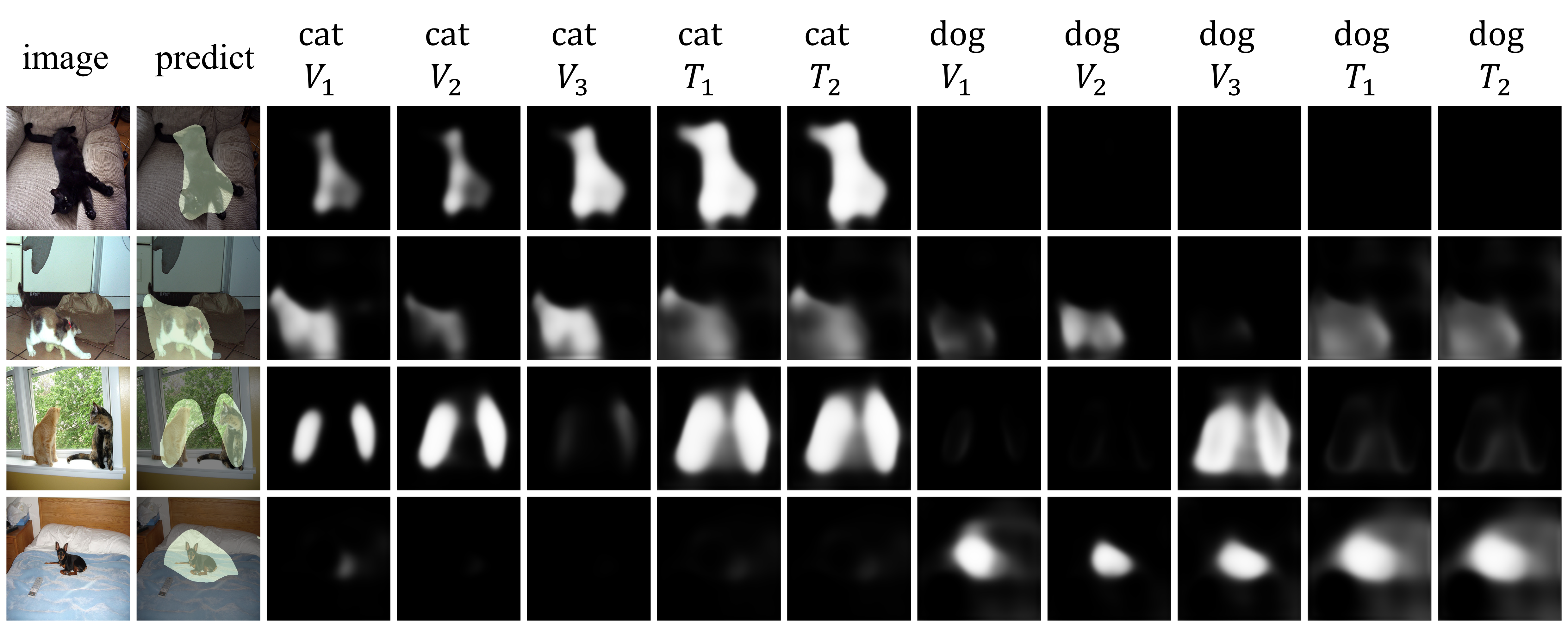}
    \caption{{\textbf{How each prototype contributes to the final prediction.} We visualize the intermediate mask by a single prototype from class ``cat'' and ``dog''. $V_1,V_2,V_3$ represents visual prototypes and $T_1,T_2$ are textual ones. Although two classes are visually similar, prototypes for the right class have higher similarity and thus dominate the prediction.}}
    \label{fig:proto_vis}
\end{figure*}
\subsection{Visualizations}
As shown in  Fig.~\ref{fig:vis}, we present some visualization results from PASCAL-$5^i$ under 5-shot setting. The classes \texttt{bus, car, cat, chair, cow} are treated as unseen. We can see that our model can segment both seen and unseen well at the same time. To see how single prototypes contribute to the corresponding class independently, we visualize the intermediate masks drawn by a single prototype as well as the final prediction in Fig.~\ref{fig:proto_vis}. 
We choose the class ``cat'' and ``dog'' for visualization, since they have different semantics, but still share some common visual features (both are furry animals). 
For each class, we choose 5 prototypes for visualization, where $V_1,V_2,V_3$ are generated by visual prototypes and $T_1,T_2$ are generated by textual prototypes.
Since the two classes are visually similar, prototypes of the other class may be wrongly activated. For example, in the third row, ``dog $V_3$'' draws attention to where there's actually a cat.
However, as can be seen in Fig.~\ref{fig:proto_vis}, prototypes for the right class have higher similarity and thus dominate the final prediction.

\section{Conclusion}
To conclude, this paper presents a novel prototype based approach to for open-world segmentation. 
Our framework leverages the complementary nature of these cues to construct powerful multi-modal prototypes, improving segmentation performance. 
To foster modality fusion, we introduce a fine-grained multi-prototype generation and fusion mechanism that efficiently merge the information of textual modality and visual modality.
The proposed method achieves state-of-the-art results on both PASCAL-$5^i$ and COCO-$20^i$ datasets.
We hope this work can motivate researchers to utilize multi-modal information for more effective and comprehensive algorithm design in the future.

\section{Acknowledgement}
This work is supported by the National Key R\&D Program of China (No. 2022ZD0160702),  STCSM (No. 22511106101, No. 22511105700, No. 21DZ1100100), 111 plan (No. BP0719010) and National Natural Science Foundation of China (No. 62306178).

\clearpage
\clearpage

\bibliography{sn-bibliography}

\end{document}